# Improving Performance of Automated Essay Scoring by using back-translation essays and adjusted scores

You-Jin Jong[1], Yong-Jin Kim[2], Ok-Chol Ri[1]

[1] Kum Sung Middle School Number 2, Pyongyang, 999093, D.P.R of Korea
[2] Faculty of Mathematics, KIM IL SUNG University, Pyongyang, 999093, D.P.R of Korea

Correspondence should be addressed to Yong-Jin Kim: kyj0916@126.com

**Abstract**: Automated essay scoring plays an important role in judging students' language abilities in education. Traditional approaches use handcrafted features to score and are time-consuming and complicated. Recently, neural network approaches have improved performance without any feature engineering. Unlike other natural language processing tasks, only a small number of datasets are publicly available for automated essay scoring, and the size of the dataset is not sufficiently large. Considering that the performance of a neural network is closely related to the size of the dataset, the lack of data limits the performance improvement of the automated essay scoring model. In this paper, we proposed a method to increase the number of essay-score pairs using back-translation and score adjustment and applied it to the Automated Student Assessment Prize dataset for augmentation. We evaluated the effectiveness of the augmented data using models from prior work. In addition, performance was evaluated in a model using long short-term memory, which is widely used for automated essay scoring. The performance of the models was improved by using augmented data to train the models.

## 1. Introduction

Currently, online education systems are being used more actively due to the COVID 19 outbreak, and the role of educational assessment systems has become more important. Learning a foreign language has become more common. Learning a foreign language is not only to satisfy our interests and flirts. In today's multicultural society, it has become essential to freely speak a second or third language. Writing is an important part of language learning. The assessment of writing ability is included in all language tests.

Automated Essay Scoring (AES) is the task of evaluating one's writing ability and assigning a score to an essay without human interference. The process of manually scoring essays is complex and time-consuming. Even though there is a fixed scoring guide, the scoring process is influenced by individual factors, such as mood and personality, and assigned scores are subjective and lack credibility. Automatic scoring for the essay was proposed as a solution to manual scoring.

All early works were based on feature engineering, and the performance was improved by adding more complicated features. In recent years, neural network models have been introduced in this research and have improved the performance without any feature engineering. Various neural networks have been used for AES, and their performance has been continuously improved. From the simplest Recurrent Neural Network (RNN) [1] and Convolutional Neural Network (CNN) to a complicated large-scale natural language pretraining model (XLNet) [2], almost every neural network has been used for AES. Prior works have improved performance by changing the structure of the neural network model or adding other features. However, we improved the performance by generating more useful data from the original data. Data augmentation techniques have been applied to other natural language processing (NLP) tasks and have shown good performance. However, there are no examples of data augmentation techniques applied to AES. Our study is the first attempt to augment AES data. We generated back-translation essays using Google Translator (https://translate.google.com/) and adjusted the corresponding scores in several ways. We trained and validated the model with doubled number of essay-score pairs and tested it on the original data. The performance is improved by using augmented data.

Our main contributions are as follows.

1. Data augmentation was introduced into AES. We proved the possibility of data augmentation by adjusting the score along with the essay.

2. We analyzed the characteristics of back-translation essays and came up with a score adjustment method suitable for back-translation essays in AES data.

3. We generated back-translation essays (English-Chinese-English, English-French- English) (https://github.com/j-y-j-109/asap-back-translation) for the Automated Student Assessment Prize (ASAP) dataset (https://www.kaggle.com/c/asap-aes).

## 2. Related Work

The first AES system was created in 1966 and uses some linguistic features to score essays [3]. Most recent works have used neural network models for AES. In 2016, [4] designed a neural network model using CNN and Long Short-Term Memory (LSTM) [11], and showed significant improvement compared to traditional methods that depend on manual feature engineering. The convolution layer extracts local features from the essay and the recurrent layer generates a representation for an essay. This is the simplest neural network model, and it generates a representation of the input essay and obtains a value from it. Since then, an increasing number of neural network models have been used for AES, and almost all neural networks including GRU [12], BERT [13], XLNet have been used (see Table 1). RNN was also used, but it was not used in the final model because the performance was lower than GRU or LSTM.

After simple models, such as [4], models that assign scores by capturing some other features, such as coherence and relevance, have also been used. Example essays were also used to assign a score to the input essay. In [8] and [10], the relevance between the prompt and the essay and the coherence of sentences within the essay were captured. In [7], the similarity between the word distribution of the input essay and that of the example essays was used. In [8], the k-means algorithm was used to classify example essays, and representative essays were selected from each cluster and used to assign scores.

Table 1: Various neural network models used for AES.

|      | CNN | GRU | LSTM | Bi-LSTM | BERT | XLNet |
|------|-----|-----|------|---------|------|-------|
| [4]  | √   |     | √    |         |      |       |
| [5]  |     |     |      | √       |      |       |
| [6]  | √   |     | √    |         |      |       |
| [7]  |     | √   |      |         |      | √     |
| [8]  |     |     | √    |         |      |       |
| [9]  |     |     |      |         | √    |       |
| [10] |     |     | √    |         |      |       |

Various models and embeddings have been used as representations for words and sentences. In [4], the pre-trained word embeddings released in [14] were used. In [5], C&W embeddings were used ([15, 16]) and augmented by considering the contribution of each word to the essay score. In [6], character embeddings were attempted, and in [6] and [7], they used pre-trained Glove embeddings trained on Google News [17]. In [8] and [10], BERT was used to obtain sentence representations.

Data augmentation techniques are widely used in NLP tasks. For the Text Classification task, [18] used the method of finding synonyms in WordNet and replacing words, and for the Categorization task, [19] obtained synonyms by calculating cosine similarity. [20] generated augmented data by translating sentences into French and again into English for Reading Comprehension, and [21] generated back-translation data using Japanese for Automatic Short Answer Scoring. To date, various data augmentation techniques have been proposed and used for many NLP tasks. However, no prior works have applied data augmentation to AES.

Back-translation means that the original data are translated into other languages and then translated back to obtain new data in the original language. This method rewrites the entire text without replacing individual words. [20] and [22] used the English-French translation model to perform back-translation for each sentence. In addition to the trained machine translation model, Google's Cloud Translation API service has been widely used ([23, 24]). [21] used the Baidu Translation API service. There are also other methods to add various additional features based on back-translation.

## 3. Augmented data

This section describes the original data and augmented data in detail. Dataset for AES consists of essays and corresponding scores. Therefore, when creating new data using data augmentation techniques, essays and corresponding scores should be determined together.

### 1) Original dataset

There are several open datasets for AES, and more than 90% of prior works were evaluated using the ASAP dataset [25]. In 2012, Kaggle hosted the ASAP competition to evaluate the capabilities of AES systems. The ASAP dataset is built with essays written by students ranging in grade levels from Grade 7 to Grade 10. There are approximately 13,000 essays corresponding to 8 prompts. For individual prompts, the number of essays is less than 2,000. Specific dataset information is presented in Table 2. Each prompt has a different score range and number of essays. The test set used in the competition is not publicly available.

Table 2: Statistics of ASAP dataset.

| Prompt | Number of Essays | Score Range |
|--------|------------------|-------------|
| 1 | 1783 | 2-12 |
| 2 | 1800 | 1-6 |
| 3 | 1726 | 0-3 |
| 4 | 1772 | 0-3 |
| 5 | 1805 | 0-4 |
| 6 | 1800 | 0-4 |
| 7 | 1569 | 0-30 |
| 8 | 723 | 0-60 |

Identifying information from the essays of the ASAP dataset was removed using the Named Entity Recognizer from the Stanford Natural Language Processing group and a variety of other approaches. The relevant entities were identified in the essay and then replaced with a string starting with '@'. Any misspelled words or grammatical errors were transcribed exactly as they occur in the original essays.

### 2) Back-translation

First, we need to obtain back-translation essays using essays of the original data. To study the general effect of back-translation essays, we generated back-translation essays using two languages. For the diversity of back-translation essays, we used multibyte language, Chinese, and single-byte language, French. As the amount of data that Google Translator can process at one time is limited, the original essays were divided into 8 equal-sized parts for translation. Google Translator perfectly translated special words starting with @ in essays.

### 3) Score adjustment

After obtaining the back-translation essays, the corresponding scores must be determined. The score setting directly affects the performance of data augmentation based on back-translation. This is because even if the number of essays in the data is large, the performance of the model can be further degraded if the scores for essays are not reasonably determined (see section 4). Therefore, it is essential to adjust the scores for back-translation essays.
The most intuitive way to set the score is to give the score of the original essay since back-translation essays are similar to original essays. In this case, the new scores are given by Equation 1:

$$s_{new}^d = s_{ori}^d, d \in E_i, i = \overline{1..8} \qquad (1)$$

where $E_i$ represents all essays of prompt $i$, $s_{ori}^d$ and $s_{new}^d$ are the original and new score of essay $d$ respectively. Another method provides a more suitable score by finely adjusting the original score. In this case, the new scores are given by Equation 2 or 3:

$$s_{new}^d = \min(s_{max}^i, s_{ori}^d + v), d \in \{e|P(e) = 1 \cap e \in E_i\}, i = \overline{1..8} \qquad (2)$$

or

$$s_{new}^d = \max(s_{min}^i, s_{ori}^d - v), d \in \{e|P(e) = 1 \cap e \in E_i\}, i = \overline{1..8} \qquad (3)$$

where $P$ is a condition. For example, let $P$ be the condition for judging whether an essay has a length greater than 300. $P(e) = 1$ means that the length of essay $e$ is greater than 300. $s_{max}^i$ and $s_{min}^i$ means the maximum and minimum scores that an essay in prompt $i$ can take. v is an additional value to adjust the score.

The essays in the ASAP dataset have certain characteristics. In the essays of the ASAP dataset, certain errors, such as grammatical and lexical errors, exist because of the characteristics of the AES task. For example, as described in section 3.1, there are some misspelled words in the ASAP dataset, and the number of misspelled words decreases after back-translation using Chinese (see Figure 1). If you use a translator, the translator can correct these errors to a certain extent in the process of translating them into other languages, and you can generate translated essays with a smaller number of errors (see Figure 2). When generating back-translation essays using these translated essays, the translator can generate essays at a relatively higher level than the original essays. Therefore, it can be assumed that the quality of back-translation essays is slightly higher than that of original essays.

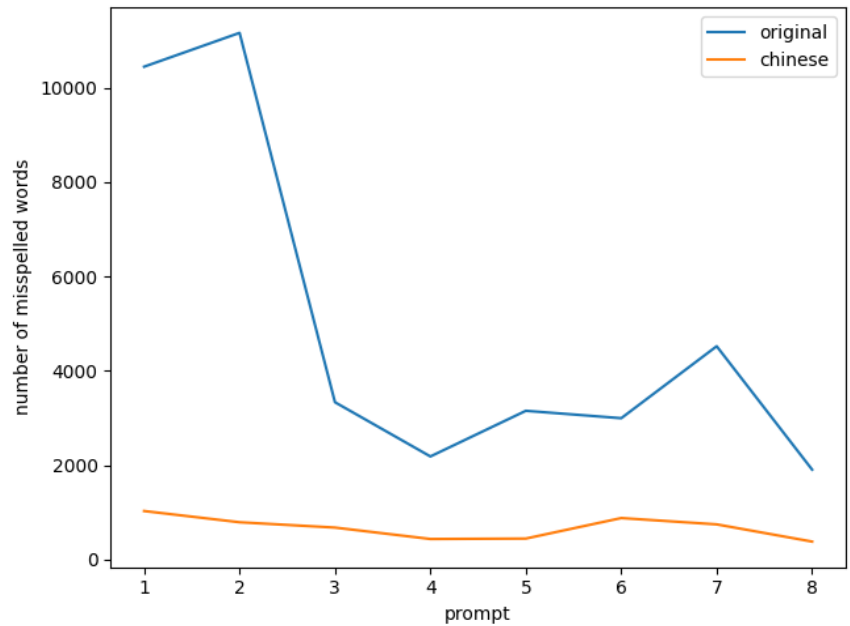

Figure 1: Number of words (the words are separated by word tokenizer of the Natural Language Tool Kit) undefined in Glove for each prompt.

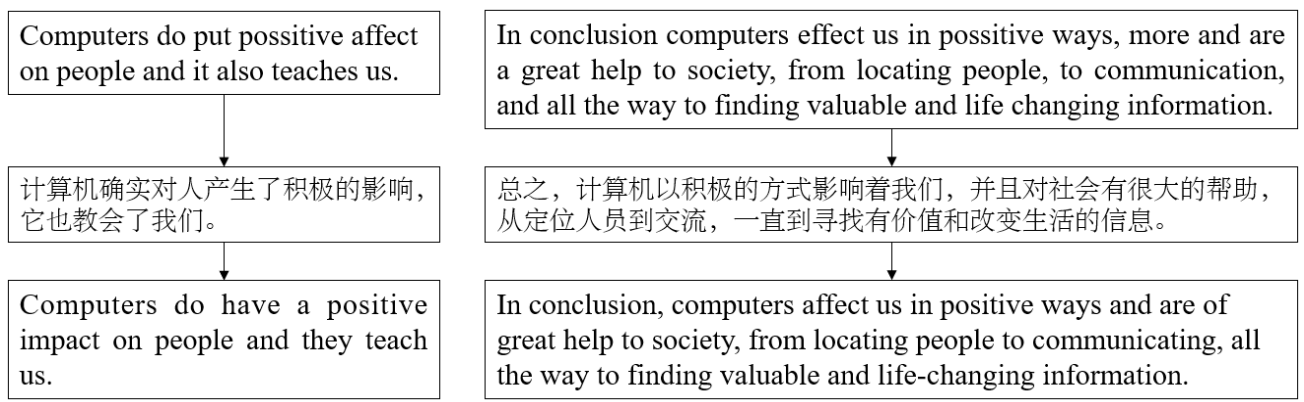

Figure 2: Results of back-translation for sentences from the ASAP dataset (essay_id: 114 and 141).

Each prompt in the ASAP dataset has a different score range (see Figure 3 and Table 3).

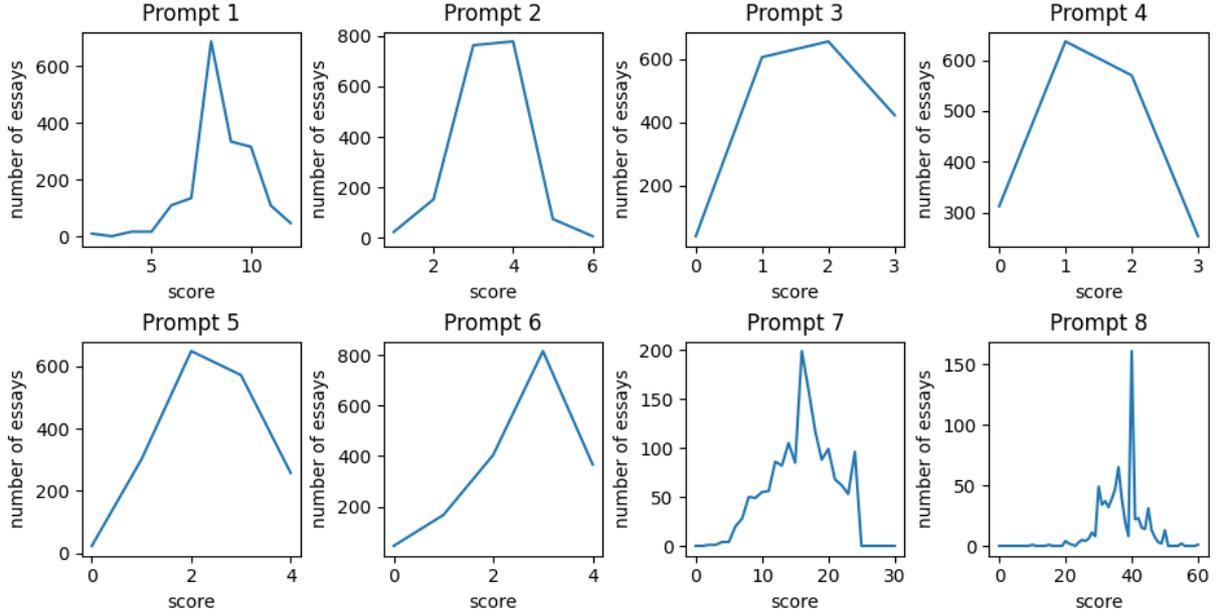

Figure 3: Score distribution for each prompt.

Table 3: Highest frequency score and number of lower/higher score essays for each prompt.

| Prompt | Score Range | Highest frequency Score | Number of lower score essays | Number of higher score essays |
|---|---|---|---|---|
| 1 | 2-12 | 8 | 977 | 806 |
| 2 | 1-6 | 4 | 1718 | 82 |
| 3 | 0-3 | 2 | 1303 | 423 |
| 4 | 0-3 | 1 | 949 | 823 |
| 5 | 0-4 | 2 | 975 | 830 |
| 6 | 0-4 | 3 | 1433 | 367 |
| 7 | 0-30 | 16 | 825 | 744 |
| 8 | 0-40 | 40 | 577 | 146 |

Considering that back-translation essays are similar to the original essays, scores of back-translation essays were given as original scores according to Equation 1 for all prompts. For prompts with a small score range, that is, 1-6, back-translation cannot change the score of essays. For example, if there are only two scores, 0 and 1, an essay with a score of 0 cannot be back-translated to a score of 1, and an essay with a score of 1 cannot be back-translated to a score of 0. A change of 1 point requires large changes in the essays. For prompts 7 and 8, Equation 2 was also used to determine the scores of back-translation essays. Condition P returns 1 when the essay has a score higher than the highest frequency score, and v is given as 1. For prompts 7 and 8, the scores of back-translation essays are given by Equation 4:

$$s_{new}^d = \min(30, s_{ori}^d + 1), d \in \{e | s_{ori}^e > 16 \cap e \in E_7\}$$

$$s_{new}^d = \min(60, s_{ori}^d + 1), d \in \{e | s_{ori}^e > 40 \cap e \in E_8\}$$

(4)

First, the additional point is set to 1 because back-translation slightly raises the quality of the essay. Second, when scoring, a higher score is usually given for a more perfectly written essay than the baseline. If you want to obtain a high score, you have to complete it more precisely in all aspects, such as vocabulary and grammar. The baseline can be assumed as the level of the essay with the highest frequency score. For high-score essays, even if there is a slight improvement, the score increases. In other words, the scores of back-translation essays from low-score essays do not increase even if back-translation is performed, but the scores of back-translation essays from high-score essays do increase after back-translation. For each prompt, the highest frequency score was found, and additional points were given to essays with a score higher than that score.

## 4. Evaluation

The models proposed in [7] were used. They countered the influence of essay length and considered the characteristics of the dataset. To evaluate the effectiveness of the augmented data and for a fair comparison, we trained the models using their published code (https://github.com/sdeva14/sustai21-counter-neural-essay-length). The effectiveness of data augmentation was evaluated by training the model using the original data and augmented data. Performance was evaluated with Quadratic Weighted Kappa (QWK). The QWK was the official criterion for the ASAP competition and was used to evaluate and compare the performance of models in many works.

### 1) Model

We used two models to determine whether the augmented data improves the model's performance (see Figure 4). As the first model, we use 'Manipulating-Length-GRU'. This model does not divide the output of the recurrent layer by the length of the essay, as in the model in [4] or other models, but by the average length of the essays included in each prompt. GRU was used as the recurrent layer.
As the second model, 'Considering-Content-LSTM' was used. This model computes the KL divergence between the word distribution of the example essays divided into three levels and the word distribution of the input essay and concatenates them to the averaged output of the recurrent layer. In this model, the output of the recurrent layer is processed as in the first model. In [7], GRU and XLNet were used, but we used LSTM, which is widely used for AES.

As a word vector, these models used Glove, a 100-dimensional pre-trained embedding model trained on Google News.

### 2) Experimental Setup

The ASAP dataset does not have a test set, and cross-validation is used to evaluate the models. We used the same cross-validation partitions as those in [4]. We trained and validated the model with doubled number of essay-score pairs and evaluated performance on the original test set. We performed 50 epochs on the validation set and applied the best model to the test set. ADAM optimizer (eps=1e-7) was used with a learning rate of 0.001. The batch size was set to 32. The cell size of the recurrent layer was 300.

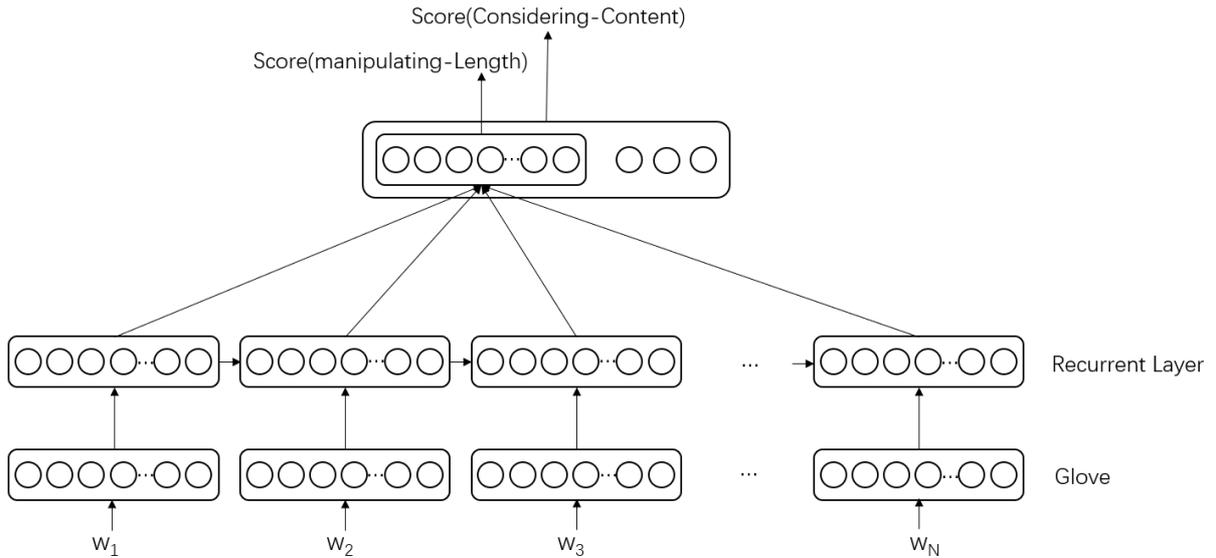

Figure 4: Architecture of the models we used.

**3) Analysis**

First, we trained the 'Manipulating-Length-GRU' model (see Table 4 and Table 5).

Table 4: Performance of 'Manipulating-Length-GRU' model for prompts 1-6.

| Data | Prompt | | | | | |
|---|---|---|---|---|---|---|
| | 1 | 2 | 3 | 4 | 5 | 6 |
| Ori | **83.7** | 69.5 | 68.7 | 80.1 | **81.5** | 80.2 |
| Ori + Ch | 83.6 | 69.5 | 68.8 | 80.4 | 81.4 | 80.5 |
| Ori + Fr | 83.5 | **69.9** | **69.0** | **80.5** | **81.5** | **80.7** |

Table 5: Performance of 'Manipulating-Length-GRU' model for prompts 7 and 8.

| Data | Prompt | |
|---|---|---|
| | 7 | 8 |
| Ori | **80.7** | 70.5 |
| Ori + Ch | 80.2 | 70.4 |
| Ori + Ch [(2), v=1] | 80.0 | 70.3 |
| Ori + Ch [(3), v=1] | 79.7 | **70.7^** |
| Ori + Ch [(2), v=2] | - | 69.7 |
| Ori + Ch [(4)] | 80.3 | 70.6 |
| Ori + Ch [(4), (5)] | - | **70.7*** |
| Ori + Fr | 80.1 | 69.9 |
| Ori + Fr [(2), v=1] | 80.1 | 70.2 |
| Ori + Fr [(3), v=1] | 79.4 | 69.8 |
| Ori + Fr [(2), v=2] | - | 69.7 |
| Ori + Fr [(4)] | 80.4 | 70.4 |

Table 4 shows the performance of the 'Manipulating-Length-GRU' model for prompts with a small score range. 'Ori' means original data. 'Ori + Ch' means augmented data with back-translation essays using Chinese. The same goes for 'Ori + Fr'. The scores of back-translation essays are the same as those of the original essays. Table 5 shows the performance of the 'Manipulating-Length-GRU' model for prompts 7, 8 with a large score range. For the augmented data using adjusted scores, we marked the score adjustment Equation number and the value of variable v in the Equation next to the data. For example, [(2), v=2] means giving all back-translation essays in the prompt 2 points higher scores than the original essay scores. The value marked with '*' is slightly bigger than the value marked with '^'.

For prompts 2, 3, 4, 6, 'Ori + Fr' showed the best performance. Except for the 'Ori + Ch' for prompt 5, for prompts 2 to 6, the performance was improved by using back-translation essays and original scores. For prompt 1, the performance did not improve, and we suspect that this is because prompt 1 has a relatively larger score range than prompts 2 to 6. For prompt 8, since the score range is twice that of prompt 7, [(2), v=2] was also applied. Except for the 'Ori + Ch [(3), v=1]' for prompt 8, if all essays of the prompt are scored using Equation 2 or Equation 3, the performance is lower than that when the score is not adjusted. In contrast, the augmented data using Equation 4 showed a higher performance than when using original scores. For prompt 8, the augmented data improved the performance compared to the original data. For prompt 7, the performance of the original data is the best.

For prompt 8, 'Ori + Ch [(3), v=1]' performed better than Ori + Ch [(4)]. Equation 4 was applied to 146 essays, as shown in Table 3, but Equation 3 was applied to a total of 723 essays. We defined Equation 5 as follows and augmented the data by using Equations 4 and 5:

$$s_{new}^d = \max\left(0, s_{ori}^d - 1\right), d \in \{e | s_{ori}^e \leq 40 \cap e \in E_8\} \tag{5}$$

The performance of the new augmented data was slightly higher than that of 'Ori + Ch [(3), v=1]'. This indicates that Equation 4 is still effective. Ori + Ch [(4)] performed worse than 'Ori + Ch [(3), v=1]' because the number of applied essays was smaller.

For prompt 5, the performance is decreased when 'Ori + Ch' is applied, and for prompt 8, the performance is improved when 'Ori + Ch [(3), v=1]' is applied. We suspect that some information is lost when translating low-score essays using a multibyte language.

Using the augmented data is twice as long as using the original data. To reduce the training time, we attempted to reduce the number of epochs for augmented data. After obtaining the result by setting the number of epochs to 50, we determined the first epoch number with the best performance (see Figure 5). When using the augmented data, the first epoch number of the best model was much smaller than when using the original data. This implies that using augmented data converges to the best model faster though increasing the training time for one epoch. Therefore, when training the next model, we set the number of epochs to 30 for augmented data. The training time of the augmented data is 1.25 times longer than that of the original data.

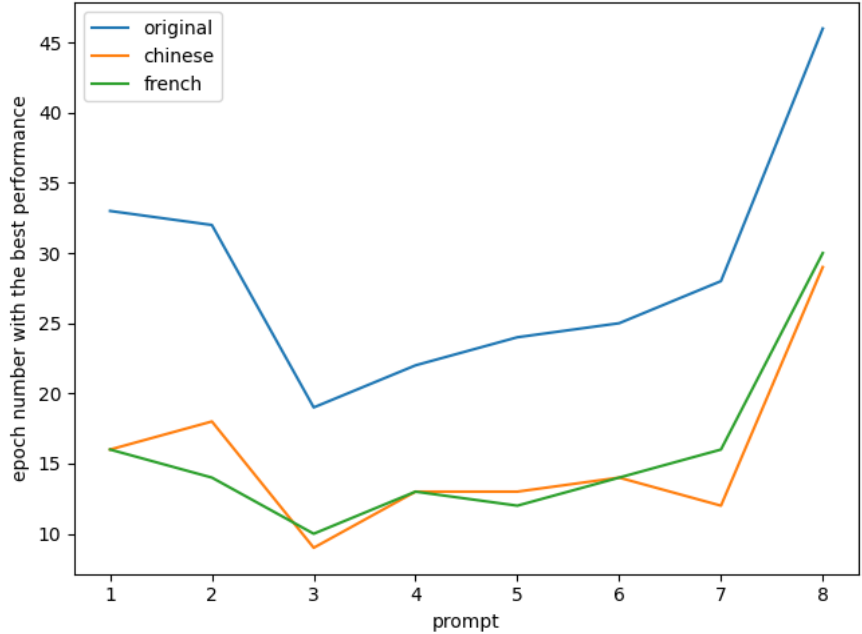

Figure 5: The first epoch number with the best performance for all prompts.

Second, we trained the 'Considering-Content-LSTM' model (see Table 6). In Table 6, the average improvement value was obtained by dividing the improvement value for 5 prompts by 8. The value marked with '*' is slightly bigger than the value marked with '^'.

Table 6: Performance of 'Considering-Content-LSTM' model for prompts 2, 3, 4, 6, and 8.

| Data | Prompt | | | | | Average |
| --- | --- | --- | --- | --- | --- | --- |
| | 2 | 3 | 4 | 6 | 8 | Improvement |
| Ori | 71.0 | 69.5 | 80.1 | 81.3 | 71.0 | - |
| Ori + Ch [(4), (5)] | 71.2 | **69.7** | 80.3 | 81.5 | **71.7** | 0.2^ |
| Ori + Fr [(4)] | **71.3** | 69.6 | **80.5** | **81.8** | 71.4 | 0.2* |

Through the experiment on the first model, for prompts 2, 3, 4, 6, and 8, the effectiveness of the augmented data was confirmed. New results were obtained using augmented data in the second model. The performance was also improved in the second model.

The performance was improved by 0.2% on average for both models using the augmented data.

## 5. Conclusions

In this paper, we improved the performance of AES by using back-translation essays and adjusted scores. We generated back-translation essays and adjust scores for the ASAP dataset, and confirmed the effectiveness of the augmented data. We used different score adjustment methods for specific prompts to find a reasonable method.

We generated back-translation essays for the ASAP dataset using Chinese and French. It was effective to maintain the score as it was for prompts with a small score range. For prompts with a large score range, based on the highest frequency score, it was effective to increase the score for the high-score essays and maintain the score for the low-score essays. For prompts 2, 3, 4, 6, and 8, higher performance was obtained than when using the original data. The performance was improved by 0.2% on average. In addition, we found that the augmented data converges to the best model faster than the original data, reducing the effect of increasing time from data augmentation to some extent.

By improving the performance of AES using data augmentation, it is possible to further improve the performance to a certain extent even when the dataset cannot be sufficiently established due to various limitations. In other words, it provided new research possibilities for the AES task, which has been conducted only through neural network model updates.

In future work, we will explore more mathematically theoretical and practical score adjustment methods for back-translation essays.

# 6. Data Availability

The ASAP dataset used in this study is available at https://www.kaggle.com/c/asap-aes. The Python codes and back-translation data used to support the findings of this study are available at https://github.com/j-y-j-109/asap-back-translation and https://github.com/sdeva14/sustai21-counter-neural-essay-length.

# 7. Acknowledgments

We thank endorsements, associate editor, and Editor-in-Chief for their valuable feedback on the paper.